\documentclass{article}

\usepackage[preprint]{neurips_2024}

\usepackage{graphicx}
\graphicspath{ {./images/} }

\usepackage{float}

\usepackage{booktabs}

\usepackage{amsmath}
\usepackage[utf8]{inputenc} 
\usepackage[T1]{fontenc}    
\usepackage{hyperref}       
\usepackage{url}            
\usepackage{booktabs}       
\usepackage{amsfonts}       
\usepackage{nicefrac}       
\usepackage{microtype}      
\usepackage{xcolor}         

\title{Microplastic Identification Using AI-Driven Image Segmentation and GAN-Generated Ecological Context}

\author{
  Alex Dils \\
  Sequoia High School \\
  \texttt{alexarthurdils@gmail.com} \\
  \And
  David Raymond \\
  Sequoia High School \\
  \texttt{davidraymond1081@gmail.com} \\
  \And
  Jack Spottiswood \\
  Sequoia High School \\
  \texttt{813667@seq.org} \\
  \And
  Samay Kodige \\
  Sequoia High School \\
  \texttt{samayk333@gmail.com} \\
  \And
  Dylan Karmin \\
  Sequoia High School \\
  \texttt{813518@seq.org} \\
  \And
  Rikhil Kokal \\
  Sequoia High School \\
  \texttt{rikhilkokal@gmail.com} \\
  \And
  Dr. Win Cowger \\
  University of California Riverside \\
  Moore Institute for Plastic Pollution Research \\
  \texttt{wcowg001@ucr.edu}\\
  \And
  Chris Sad\'ee \\
  Stanford University \\
  Center for Biomedical Informatics Research \\
  \texttt{sadee@stanford.edu}
}

\begin{document}

\maketitle

\begin{abstract}
  Current methods for microplastic identification in water samples are costly and require expert analysis. Here, we propose a deep learning segmentation model to automatically identify microplastics in microscopic images. We labeled images of microplastic from the Moore Institute for Plastic Pollution Research and employ a Generative Adversarial Network (GAN) to supplement and generate diverse training data. To verify the validity of the generated data, we conducted a reader study where an expert was able to discern the generated microplastic from real microplastic at a rate of 68\%. Our segmentation model trained on the combined data achieved an F1-Score of 0.91 on a diverse dataset, compared to the model without generated data's 0.82. With our findings we aim to enhance the ability of both experts and citizens to detect microplastic across diverse ecological contexts, thereby improving the cost and accessibility of microplastic analysis.
\end{abstract}

\section{Introduction}

Microplastic pollution poses a significant threat to the health of marine ecosystems and safe drinking water [1]. Therefore, the ability to quickly, cheaply, and accurately analyze samples becomes an invaluable tool for scientists and public health officials worldwide. Large-scale microplastic analysis is currently hindered by a lack of cheap and accessible measuring equipment [2]. Microplastic analysis requires long laboratory-intensive physical and chemical pretreatments to prepare the sample. This workup often entails a combination of manual analysis and Fourier-Transform Infrared Spectrophotometers to measure microplastic material type and confidently identify microplastics [3]. Manual particle picking using visual microscopy is error-prone due to some natural particles looking similar to microplastics and vice versa. The automated approach is limited by other pollutants that may interact with light in a similar way to microplastic [4]. Both are limited by their cost, which restricts access to only well-funded research institutions or organizations. To gain a more comprehensive and widespread understanding of microplastic pollution in different environments, it's critical to provide an accurate but accessible method of measuring it.

Deep learning segmentation models have shown immense potential in accurately identifying and quantifying various elements within images while being relatively inexpensive compared to traditional methods. For example, in radiology, segmentation models are used to identify tumors in CT scan images with high precision [5]. Here, we propose a deep learning segmentation model for microplastic detection that offers a cost-effective, efficient, and scalable solution for analyzing microplastics [6]. With an effective segmentation model, scientists can process vast datasets of images from various sources, automatically identifying and quantifying microplastics.

Deep learning models such as the one we propose require diverse training data. Unfortunately, many existing microscopic images of microplastic fit for computer vision models are isolated---focusing solely on the microplastic in the image while all other pollutants are removed during pretreatment. Due to this, there’s a lack of training data on microplastics in diverse ecological scenarios, e.g., dirty samples or raw imagery of plastic in the environment [7]. Therefore, the presence of other particles present in diverse ecological environments may pose an issue for a segmentation model optimized largely on isolated microplastic samples. To combat this lack of data from ecological environments, we propose an inpainting generative adversarial network (GAN) to introduce artificial training data of microplastic in images of diverse ecological environments to train our segmentation model [8].

\section{Literature Review}

Recent advancements in artificial intelligence (AI) have significantly influenced environmental monitoring, particularly in the analysis of microplastics. Zhang et al. (2023) provide a comprehensive review of AI-based microplastic imaging technologies. AI methodologies, especially deep learning models like convolutional neural networks (CNNs), have proven effective in automating the identification and quantification of microplastics in diverse environments [9]. However, Zhang et al. emphasize the importance of developing robust datasets to support AI model training, which is critical for advancing these methods further. For example, AI integration with technologies like Unmanned Aerial Vehicles (UAVs) has the potential to expand monitoring capabilities over large areas, providing a cost-effective solution for identifying microplastics [10][11].

According to Zhang et al. (2023), although deep learning models have greatly improved the efficiency and accuracy of microplastic analysis, these models often require extensive labeled datasets to train effectively [12]. This requirement poses a challenge, as many existing datasets are fragmented and not readily accessible due to limited data-sharing practices among research institutions [9][13]. Previous studies, such as those by Geyer et al. (2017) and Hidalgo-Ruz et al. (2012), have also underscored the global impact of microplastic pollution and the urgent need for improved detection and classification methods [14][15]. Moreover, they advocate for a stronger emphasis on international collaboration to facilitate data sharing from diverse environments, which could significantly advance research in this field [9]. A step towards this goal has been demonstrated by Yurtsever et al. (2019) in their approach of using multiple different water sources and wastewaters to train a CNN to detect microbeads [16]. Approaches like these will be crucial in addressing the widespread issue of microplastic pollution and in developing more effective monitoring and mitigation strategies [10][14].

\section{Data Collection}

We collected three distinct datasets for model development:

\textbf{Cohort 1}: 368 microscopic images cleaned from pollutants other than microplastics from the "Microplastic Image Explorer" dataset provided by the Moore Institute for Plastic Pollution Research [16]. These images were used for training the GAN.

\textbf{Cohort 2}: 733 microscopic images from unlicensed sources for public use depicting various ecological scenarios. These images do not contain microplastic and, hence, were processed through the trained GAN to generate the diverse training data.

\textbf{Cohort 3}: 102 images from unlicensed public sources showing microplastics in different ecological contexts, used for evaluating the segmentation models.

For cohorts 1 and 3, we manually segmented the microplastics in each image, creating binary masks where white regions represent microplastics and black regions represent non-microplastic areas. Microplastics of morphologies including fibers and film were used for all training and testing.

\section{Model Training}

In this section, we describe the training processes for the GAN model modified to inpaint microplastic and the segmentation model. Our approach integrates the inpainting model to augment the dataset with realistic synthetic images of microplastics in various ecological environments.

\subsection{Inpainting GAN}

To generate microplastic in images using an inpainting GAN, we employed a preexisting GAN architecture consisting of a generator \( G \) and a discriminator \( D \) working in an adversarial manner [18]. The generator \( G \) learns to inpaint masked regions in the input images, while the discriminator \( D \) learns to distinguish between real images and the images generated by \( G \).

For our purposes in generating microplastic, we changed the output of the model to only include the masked region after the decoder in the generator. The binary mask guides the generator by indicating which parts of the image should be inpainted (white regions) and which parts should remain unchanged (black regions) thereby mitigating the artifacts and inconsistencies that are well-documented as limitations with the use of GANs [19].

The unmasked regions remain unaltered by combining masked and original images as follows

\begin{equation}
\text{output} = G(\text{image}, \text{mask}) \cdot \text{mask} + \text{image} \cdot (1 - \text{mask})
\end{equation}

Here, the term \( G(\text{image}, \text{mask}) \cdot \text{mask} \) ensures that only the masked areas are inpainted, while the term \( \text{image} \cdot (1 - \text{mask}) \) preserves the unmasked regions of the original image.

The model was trained on the images and masks from Cohort 1 using an NVIDIA GeForce RTX 4060, following the same training procedures as the original preexisting version.

After training the inpainting GAN model, it was applied to generate realistic synthetic images of microplastic within various ecological contexts using the images from Cohort 2. During this application, we randomly selected a mask from Cohort 1 to act as the guiding mask. The guiding mask was then randomly transformed in several ways: it was shifted vertically and horizontally, moved up and down, and rotated between 0 and 360 degrees. This transformation process seeks to add more diverse training data.

\begin{figure}[!htb]
    \centering
    \includegraphics[width=0.5\textwidth]{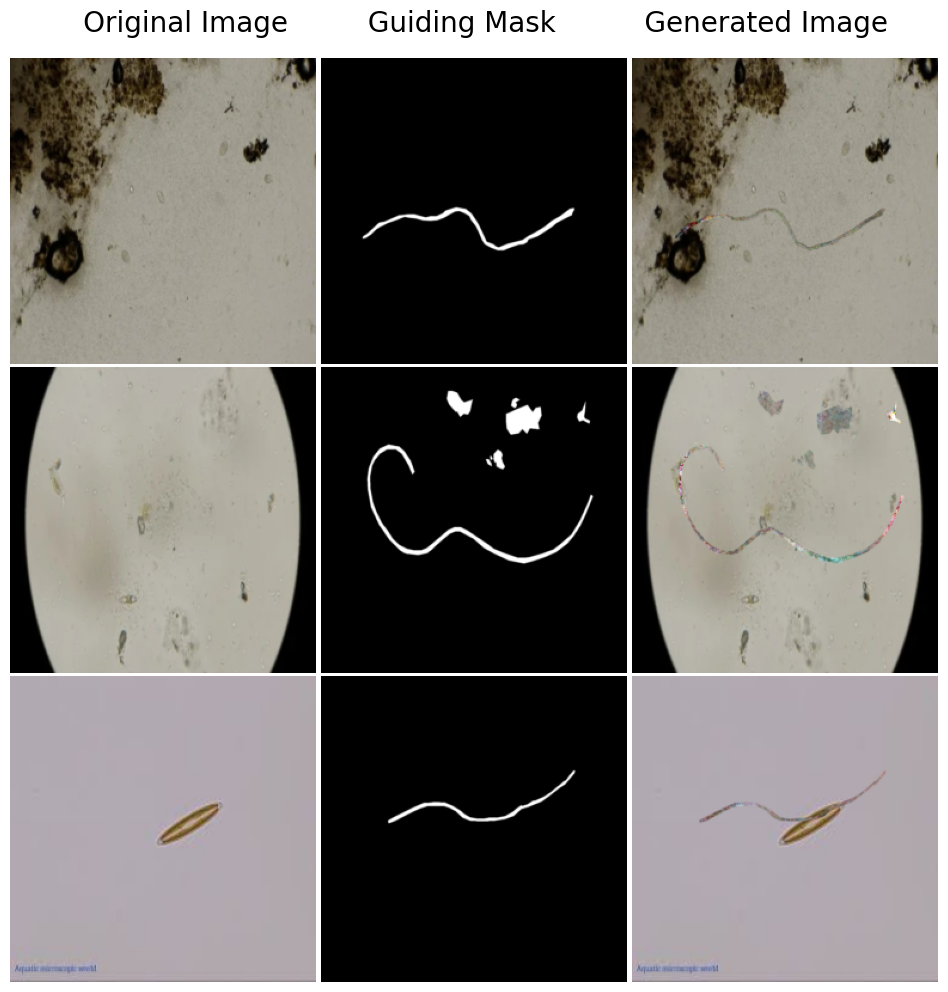}
    \caption{Comparison of original images, guiding masks, and GAN-generated images. The guiding masks indicate regions for inpainting, resulting in generated images that integrate microplastic representations within the original image context.}
    \label{fig:your_label}
\end{figure}

\subsection{Segmentation}
We trained two segmentation models. The first trained exclusively on Cohort 1, and the second trained on Cohort 1 with the additional images generated by applying the inpainting GAN to Cohort 2. Both were trained using an NVIDIA GeForce RTX 4060.

To develop the segmentation models, we used a preexisting U-Net architecture with an InceptionV3 backbone [20]. The datasets were split into training, validation, and test sets, with an 80/10/10 split. The models were trained on the train set and the parameters and data augmentation were kept constant from the preexisting implementation.

\section{Experiments}

\subsection{Inpainting GAN Performance}

To evaluate the ability of the inpainting GAN model to generate realistic synthetic images of microplastic in diverse environments, we developed a reader study. In this study, Coauthor Dr. Cowger (an expert in microplastic analysis) was blinded to the model outputs and was asked to distinguish between real and GAN-generated images of microplastic.

The study involved a randomized set of 200 images: 100 real microscopic images of microplastics taken from Cohort 3 and 100 generated by the inpainting GAN from Cohort 2. The participant was presented with these images and asked to classify each one as either real or generated. The results of the reader study indicated that GAN-generated images were correctly identified with only 68\% accuracy. This relatively low discernment rate suggests that the GAN-generated images were highly realistic and closely resembled the true images of microplastics.

\subsection{Segmentation Performance}
To evaluate the performance of the segmentation model, we measured the F1-score and Dice score for each model on Cohort 3 and the separate test set from Cohort 1. The F1-score provides a balance between precision and recall, whereas the Dice score is a measure of overlap between the predicted and true segmentation. These metrics were chosen to provide a comprehensive evaluation of the models' performance.

\begin{table}[H]
  \centering
  \begin{tabular}{ll|cc|cc}
    \toprule
    \textbf{Model} & \textbf{Training Dataset} & \multicolumn{2}{c|}{\textbf{Cohort 3}} & \multicolumn{2}{c}{\textbf{Cohort 1 Test Set}} \\
    \cmidrule(r){3-4} \cmidrule(r){5-6}
    & & \textbf{F1-Score} & \textbf{Dice Score} & \textbf{F1-Score} & \textbf{Dice Score} \\
    \midrule
    1 & Cohort 1 & 0.82 & 0.83 & 0.80 & 0.81 \\
    2 & Cohort 1 + Inpainted Cohort 2 & \textbf{0.91} & \textbf{0.92} & \textbf{0.88} & \textbf{0.89} \\
    \bottomrule
  \end{tabular}
  \caption{Performance Comparison of Segmentation Models}
  \label{performance-table}
\end{table}

Both models show satisfactory performance, but Model 2 outperforms Model 1 across all metrics and datasets. This indicates that the inclusion of GAN-generated images in the training set significantly improves the model's ability to generalize to unseen data, including microplastic in diverse ecological scenarios.

\section{Implementation}

Web-based applications make AI accessible to researchers and the public. Inspired by tools like \textit{Trash AI} (Cowger et al., 2023), which allows real-time identification of trash types, our platform allows users to upload images and receive real-time analysis without the need for specialized hardware or technical expertise [21]. The application is built using Flask for the backend to handle model inference and web requests, and a simple frontend created with HTML and CSS for user interaction [22]. The link to our website is as follows: \href{https://b446-2001-5a8-4139-7800-a50e-96a1-1b56-4a0f.ngrok-free.app}{https://b446-2001-5a8-4139-7800-a50e-96a1-1b56-4a0f.ngrok-free.app}.

Users can upload images of water samples through the web interface, which are then processed by our deep learning models to identify and segment microplastics. The segmentation results are displayed back to the user in real-time.

This implementation allows researchers and citizens to leverage advanced AI models without needing specialized hardware or software, thereby democratizing access to high-quality microplastic analysis tools.

\section{Limitations}

There are several limitations to consider with our results. First, the GAN-generated images, although realistic, may not capture the full variability and complexity of real-world ecological scenarios. The reader study indicates a 68\% discernment rate, suggesting room for improvement in the GAN's ability to replicate microplastic features accurately. Additionally, our dataset, despite being diverse, may not encompass all possible environmental contexts where microplastics occur. This could limit the generalizability of our model to new, unseen ecological environments. Lastly, the implementation of our web-based application, while aimed at democratizing access to microplastic analysis, relies on stable internet connections and may face challenges in areas with limited technological infrastructure.

\section{Conclusion}

This paper offers a novel approach to the identification of microplastics through the use of deep-learning segmentation models augmented with synthetic images generated by an inpainting GAN. The results demonstrate that the inclusion of GAN-generated images significantly improves the model’s performance, as evidenced by higher F1 scores and Dice scores on both the external test set (Cohort 3) and the test set from Cohort 1.

In conclusion, the integration of GAN-generated images into the training process represents an advancement in how microplastics can be identified in a more cheap and accessible manner. This methodology not only improves model performance but also opens up new avenues for research and application, ultimately contributing to more effective and widespread environmental monitoring and protection. This enhanced capability can facilitate more efficient and cost-effective monitoring of microplastic pollution, contributing to environmental protection efforts and public health initiatives.

\section{Acknowledgments}

We thank the Moore Institute for Plastic Pollution Research for their dataset of microplastic images, which was instrumental to our model training. Finally, we thank all members of the Coding Club at Sequoia High School for their support and advice during the project.

The code and data used in this study are publicly available. The code for the deep learning segmentation model and the inpainting GAN can be found on GitHub: \href{https://github.com/axel-slid/Microplastic-Segmentation-GAN/tree/main}{https://github.com/axel-slid/Microplastic-Segmentation-GAN/tree/main}. The datasets used for training and evaluation are available at: \href{https://dataverse.harvard.edu/dataverse/Microplastic-Segmentation-GAN/}{https://dataverse.harvard.edu/dataverse/Microplastic-Segmentation-GAN/}.

\section*{References}

\medskip

{\small

[1] Ziani, Khaled et al. (2023) Microplastics: A Real Global Threat for Environment and Food Safety: A State of the Art Review. \textit{Nutrients} \textbf{15}(3):617. doi:10.3390/nu15030617.

[2] Stock, F. et al. (2020) Pitfalls and Limitations in Microplastic Analyses. \textit{Plastics in the Aquatic Environment - Part I. The Handbook of Environmental Chemistry}, vol 111. Springer, Cham. doi:10.1007/698\_2020\_654.

[3] Primpke, S., Christiansen, S.H., Cowger, W., et al. (2020) Critical Assessment of Analytical Methods for the Harmonized and Cost-Efficient Analysis of Microplastics. \textit{Applied Spectroscopy} \textbf{74}(9):1012-1047. doi:10.1177/0003702820921465.

[4] Levine, S.P., Li-Shi, Y., Strang, C.R., \& Hong-Kui, X. (1989) Advantages and Disadvantages in the Use of Fourier Transform Infrared (FTIR) and Filter Infrared (FIR) Spectrometers for Monitoring Airborne Gases and Vapors of Industrial Hygiene Concern. \textit{Applied Industrial Hygiene} \textbf{4}(7):180-187. doi:10.1080/08828032.1989.10390419.

[5] Sabir, M.W., Khan, Z., Saad, N.M., Khan, D.M., Al-Khasawneh, M.A., Perveen, K., Qayyum, A., \& Azhar Ali, S.S. (2022) Segmentation of liver tumor in CT scan using ResU-Net. \textit{Applied Sciences} \textbf{12}(17):8650. doi:10.3390/app12178650.

[6] Iakubovskii, P. (2019) Segmentation Models. \textit{GitHub repository}. Available at: \url{https://github.com/qubvel/segmentation_models}.

[7] Li, Changchao et al. (2021) “Microplastic communities” in different environments: Differences, links, and role of diversity index in source analysis. \textit{Water Research} \textbf{188}:116574. doi:10.1016/j.watres.2020.116574.

[8] Zeng, Y., Fu, J., Chao, H., \& Guo, B. (2021) Aggregated Contextual Transformations for High-Resolution Image Inpainting. \textit{arXiv preprint arXiv:2104.01431}. doi:10.48550/arXiv.2104.01431.

[9] Zhang, Y.; Zhang, D.; Zhang, Z. A Critical Review on Artificial Intelligence—Based Microplastics Imaging Technology: Recent Advances, Hot-Spots, and Challenges. {\it Int. J. Environ. Res. Public Health} {\bf 2023}, {\bf 20}, 1150. doi:10.3390/ijerph20021150.

[10] Martin, C.; Parkes, S.; Zhang, Q.; Zhang, X.; McCabe, M.F.; Duarte, C.M. Use of unmanned aerial vehicles for efficient beach litter monitoring. {\it Mar. Pollut. Bull.} {\bf 2018}, {\bf 131}, 662–673. doi:10.1016/j.marpolbul.2018.04.040.

[11] Takaya, K.; Shibata, A.; Mizuno, Y.; Ise, T. Unmanned aerial vehicles and deep learning for assessment of anthropogenic marine debris on beaches on an island in a semi-enclosed sea in Japan. {\it Environ. Res. Commun.} {\bf 2022}, {\bf 4}, 015003. doi:10.1088/2515-7620/ac4666.

[12] Jeong, J.; Choi, J. Development of AOP relevant to microplastics based on toxicity mechanisms of chemical additives using ToxCast™ and deep learning models combined approach. {\it Environ. Int.} {\bf 2020}, {\bf 137}, 105557. doi:10.1016/j.envint.2020.105557.

[13] Brandt, J.; Mattsson, K.; Hassellöv, M. Deep Learning for Reconstructing Low-Quality FTIR and Raman Spectra–A Case Study in Microplastic Analyses. {\it Anal. Chem.} {\bf 2021}, {\bf 93}, 16360–16368. doi:10.1021/acs.analchem.1c03162.

[14] Geyer, R.; Jambeck, J.R.; Law, K.L. Production, use, and fate of all plastics ever made. {\it Sci. Adv.} {\bf 2017}, {\bf 3}, e1700782. doi:10.1126/sciadv.1700782.

[15] Hidalgo-Ruz, V.; Gutow, L.; Thompson, R.; Thiel, M. Microplastics in the Marine Environment: A Review of the Methods Used for Identification and Quantification. {\it Environ. Sci. Technol.} {\bf 2012}, {\bf 46}, 3060–3075. doi:10.1021/es2031505.

[16] Yurtsever, M., Yurtsever, U. (2019) Use of a convolutional neural network for the classification of microbeads in urban wastewater. {\it Chemosphere} {\bf 216}:271-280. doi:10.1016/j.chemosphere.2018.10.084.

[17] Sherrod, H., et al. "One4All: An Open Source Portal to Validate and Share Microplastics Data and Beyond." Journal of Open Source Software 9.99 (2024): 6715. 

[18] Liu, Ming-Yu and Tuzel, Oncel. (2016) Coupled Generative Adversarial Networks. \textit{arXiv preprint arXiv:1606.07536}. doi:10.48550/arXiv.1606.07536.

[19] Zhang, Xu, Svebor Karaman, and Shih-Fu Chang. "Detecting and simulating artifacts in GAN fake images." 2019 IEEE International Workshop on Information Forensics and Security (WIFS). IEEE, 2019.

[20] Iakubovskii, P. (2019) Segmentation Models. {\it GitHub repository}. Available at: \url{https://github.com/qubvel/segmentation_models}

[21] Cowger, W., Hollingsworth, S., Fey, D., Norris, M.C., Yu, W., Kerge, K., Haamer, K., Durante, G., \& Hernandez, B. (2023) Trash AI: A Web GUI for Serverless Computer Vision Analysis of Images of Trash. {\it Journal of Open Source Software} {\bf 8}(89):5136. doi:10.21105/joss.05136.

[22] Grinberg, Miguel. (2018) Flask web development: developing web applications with python. {\it O'Reilly Media, Inc.}

}

\end{document}